\documentclass[runningheads]{llncs}
\usepackage{graphicx}

\usepackage{tikz}
\usepackage{comment}
\usepackage{amsmath,amssymb} 
\usepackage{color}
\usepackage{wrapfig}
\usepackage{multicol}
\usepackage{makecell}
\usepackage[accsupp]{axessibility}  
\usepackage{array}

\usepackage{enumitem}
\usepackage[pagebackref,breaklinks,colorlinks]{hyperref}
\usepackage{multirow}
\usepackage{rotating}

\usepackage{marvosym}

\newcommand{\ie}{\textit{i}.\textit{e}.}
\newcommand{\eg}{\textit{e}.\textit{g}.}

\begin{document}
\pagestyle{headings}
\mainmatter
\def\ECCVSubNumber{2269}  

\title{Multi-Modal Masked Pre-Training \\ for Monocular Panoramic Depth Completion} 

\titlerunning{M$^3$PT for Monocular Panoramic Depth Completion}
\authorrunning{Yan et al.}
\author{Zhiqiang Yan$^{*}$, Xiang Li$^{*}$, Kun Wang, Zhenyu Zhang, \\Jun Li\textsuperscript{\Letter}, and Jian Yang\textsuperscript{\Letter}}
\institute{PCA Lab, Nanjing University of Science and Technology, China\\
\email{\{Yanzq,xiang.li.implus,kunwang,junli,csjyang\}@njust.edu.cn}\\
\email{zhangjesse@foxmail.com}\\
$^*$ Equal contribution
}

\maketitle
 \begin{figure}[h!]
  \centering
  \includegraphics[width=1\columnwidth]{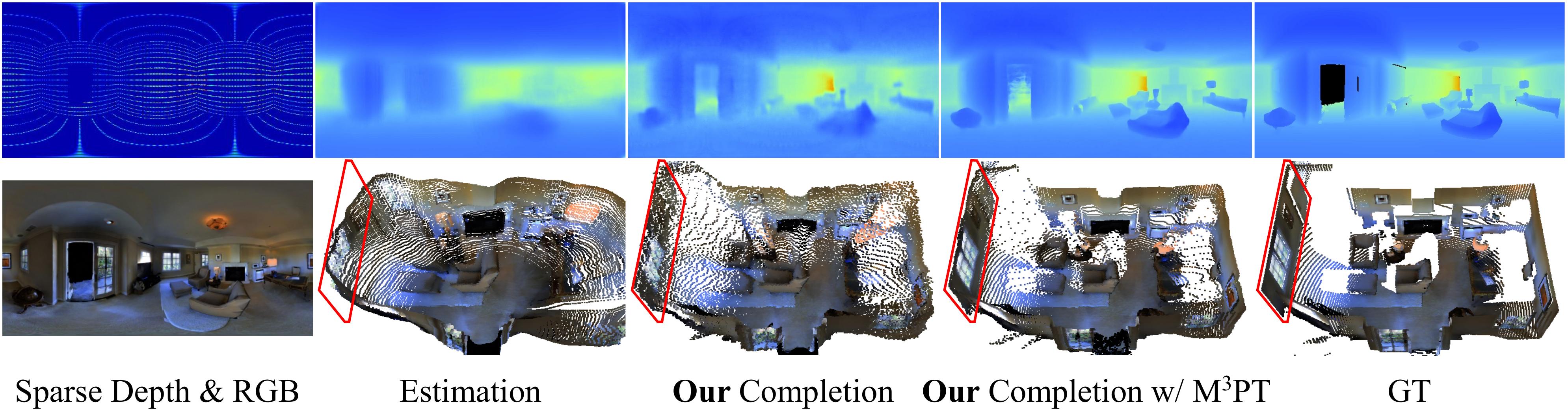}
  \caption{Comparisons of the predicted depth and 3D reconstruction results between panoramic depth estimation (with RGB) and completion (with RGB and sparse depth).}\label{dc_pretrain}
\end{figure}

\begin{abstract}
In this paper, we formulate a potentially valuable panoramic depth completion (PDC) task as panoramic 3D cameras often produce 360$^\circ$ depth with missing data in complex scenes. Its goal is to recover dense panoramic depths from raw sparse ones and panoramic RGB images. To deal with the PDC task, we train a deep network that takes both depth and image as inputs for the dense panoramic depth recovery. However, it needs to face a challenging optimization problem of the network parameters due to its non-convex objective function. To address this problem, we propose a simple yet effective approach termed M$^{3}$PT: multi-modal masked pre-training. 
Specifically, during pre-training, we simultaneously cover up patches of the panoramic RGB image and sparse depth by shared random mask, then reconstruct the sparse depth in the masked regions. To our best knowledge, it is the first time that we show the effectiveness of masked pre-training in a multi-modal vision task, instead of the single-modal task resolved by masked autoencoders (MAE). 
Different from MAE where fine-tuning completely discards the decoder part of pre-training, there is no architectural difference between the pre-training and fine-tuning stages in our M$^{3}$PT as they only differ in the prediction density, which potentially makes the transfer learning more convenient and effective. 
Extensive experiments verify the effectiveness of M$^{3}$PT on three panoramic datasets. Notably, we improve the state-of-the-art baselines by averagely 29.2\% in RMSE, 51.7\% in MRE, 49.7\% in MAE, and 37.5\% in RMSElog on three benchmark datasets. 

\end{abstract}

\section{Introduction}
\begin{figure}[t]
  \centering
  \includegraphics[width=1\columnwidth]{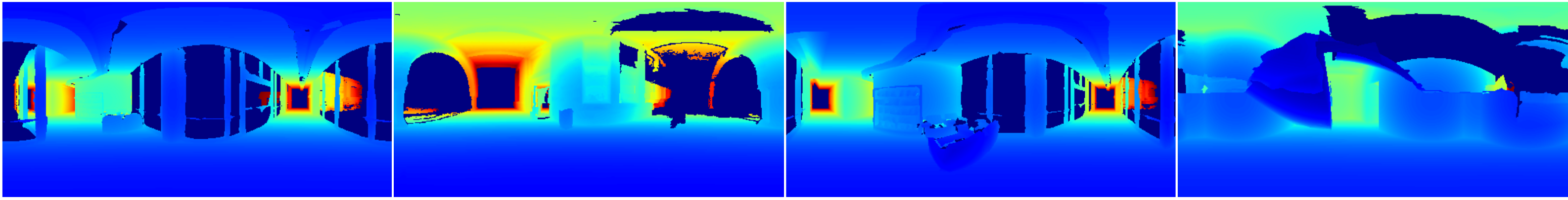}\\
  \caption{Panoramic depth maps with large missing areas shown in darkest blue color.}\label{fig:missingdata}
\end{figure}

Panoramic depth perception (see Table \ref{task}) has received increasing attention in both academic and industrial communities due to its crucial role in a wide variety of downstream applications, such as virtual reality \cite{albanis2021pano3d}, scene understanding \cite{sun2021hohonet}, and autonomous navigation \cite{gordon2019depth}. With the development of hardware devices, panoramic 3D cameras become easier and cheaper to capture both RGB and depth (RGB-D) data with 360$^\circ$ field of view (FoV). Depending on the captured RGB images, all recent perception technologies \cite{pintore2021slicenet,sun2021hohonet,jiang2021unifuse,li2021panodepth,yun2021improving,zhuang2021acdnet,bai2022glpanodepth,shen2021distortion,shen2022panoformer}, to the best of our knowledge, concentrate on panoramic depth estimation (PDE) that predicts dense depth from a single 360$^\circ$ RGB image. In this paper, we focus on exploring the 360$^\circ$ RGB-D pairs for the panorama perception with an effective pre-training strategy. We show our motivations as follows:

\textbf{Two Motivations for Panoramic Depth Completion} (PDC). \emph{One is the 360$^\circ$ depth maps with missing areas.} During the collection process, the popular panoramic 3D cameras (\eg, Matterport Pro2\footnote{https://matterport.com/cameras/pro2-3D-camera} and FARO Focus\footnote{https://www.faro.com/en/Products/Hardware/Focus-Laser-Scanners}) still produce 360$^\circ$ depth with missing areas when facing bright, glossy, thin or far surfaces, especially indoor rooms in Figure \ref{fig:missingdata}. These depth maps will result in a poor panorama perception. To overcome this problem, we consider a new panoramic depth completion task, completing the depth channel of a single 360$^\circ$ RGB-D pair captured from a panoramic 3D camera. \emph{Another is an experimental investigation that in contrast with PDE, PDC is much fitter for the panoramic depth perception.} For simplicity and fairness, we directly employ the same network architectures (\eg, UniFuse \cite{jiang2021unifuse}) to estimate or complete the depth map. Figure \ref{comparison_de_dc_M3PT} reports that the PDC has much lower root mean square error than PDE. Furthermore, Figure \ref{dc_pretrain} shows that the PDC can recover more precise 360$^\circ$ depth values, leading to better 3D reconstruction. This observation reveals that PDC is more important than PDE for 3D scene understanding.

\textbf{One Motivation for Pre-Training}. When using deep networks to perceive the depth information, there is a challenging problem: \emph{how to optimize the parameters of the deep networks}? It is well-known that the objective function is highly non-convex, resulting in many distinct local minima in the network parameter space. Although the completion can lead to better network parameters and higher accuracy on the depth perception, it is still not to satisfy practical needs of the 3D reconstruction. Here, we are inspired by the greedy layer-wise pre-training technology \cite{Erhan2010pretraining,li2016sparseness} that stacks two-layer unsupervised autoencoders to initialize the networks to a point in parameter space, and then fine-tunes them in a supervised setting. This technology drives the optimization process more effective, achieving a `good' local minimum. Recently, the single-modal masked autoencoders \cite{he2021masked,xie2021simmim} are also applied into object detection and semantic segmentation, achieving amazing improvements on their benchmarks. These interpretations and improvements motivate us to explore a new pre-training technology for the multi-modal panoramic depth completion.

In this paper, we propose a Multi-Modal Masked Pre-Training (M$^3$PT) technology to directly initialize all parameters of deep completion networks. Specifically, the key idea of M$^3$PT is to employ a shared random mask to simultaneously corrupt the RGB image and sparse depth, and then use the invisible pixels of the sparse depth as supervised signal to reconstruct the original sparse depth. After this pre-training, no-masked RGB-D pairs are fed into the pre-trained network supervised by dense ground-truth depths. Different from the layer-wise pre-training \cite{Erhan2010pretraining}, M$^3$PT is to pre-train all layers of the deep network. Compared to MAE \cite{he2021masked}, M$^3$PT has no architectural difference between the pre-training and fine-tuning stages, where they differ in only the prediction density of target depth. This characteristic probably makes it convenient and effective for the transfer learning, including but not limited to the multi-modal depth completion, denoising, and super-resolution guided by RGB images. In summary, our contributions are as follows:
\begin{itemize}[leftmargin=*]
\setlength{\itemsep}{0pt}
  \item We introduce a new panoramic depth completion (PDC) task that aims to complete the depth channel of a single 360$^\circ$ RGB-D pair captured from a panoramic 3D camera. To the best of our knowledge, we are the first to study the PDC task to facilitate 360$^\circ$ depth perception.
  \item We propose the multi-modal masked pre-training (M$^3$PT) for the multi-modal vision task. Different from the layer-wise pre-training \cite{Erhan2010pretraining} and MAE \cite{he2021masked}, M$^3$PT is to pre-train all layers of the deep network, and does not change the network architecture in the pre-training and fine-tuning stages.
  \item On three benchmarks, \ie, Matterport3D \cite{albanis2021pano3d}, Stanford2D3D \cite{armeni2017joint}, and 3D60 \cite{zioulis2019spherical}, extensive experiments demonstrate that (i) PDC achieves higher accuracy of panoramic depth perception than PDE, and (ii) our M$^3$PT technology achieves the state-of-the-art performance.
\end{itemize}

\begin{table}[t]
\centering
\Large
\renewcommand\arraystretch{1.2}
\resizebox{1\textwidth}{!}{
\begin{tabular}{l|c|c|c|c}
\Xhline{1.2pt}
Task            & Depth Estimation  & Depth Completion  & Panoramic Depth Estimation & Panoramic Depth Completion \\ \hline
New task        & No                & No                & No             & Yes           \\
Data FoV        & $<$180$^{\circ}$  & $<$180$^{\circ}$  & 360$^{\circ}$  & 360$^{\circ}$ \\
Data modal      & RGB               & RGB-D             & RGB            & RGB-D         \\
Camera          & perspective       & perspective       & panoramic      & panoramic     \\
Target          & depth             & depth             & depth \& 3D reconstruction  & depth \& 3D reconstruction \\ 
\Xhline{1.2pt}
\end{tabular}
}
\caption{Comparisons of different depth related tasks. FoV denotes the field of vision.}
\label{task}
\end{table}

\section{Related Work}
Since this paper aims to learn the new task of monocular panoramic depth completion, we report three related but different topics whose detailed differences are listed in Table~\ref{task}. First, we review depth completion approaches that input single RGB-D pair with limited FoV. Second, we elaborate on panoramic depth estimation works which predict 360$^{\circ}$ depths from panoramic color images. At last, we introduce the masked image encoding technology.

\subsection{Monocular Depth Completion with Limited FoV}
Existing monocular depth completion methods primarily focus on sparse depths and color images with a narrow FoV less than 180$^{\circ}$. Up to now, based on the commonly used KITTI \cite{Uhrig2017THREEDV} and NYUv2 \cite{silberman2012indoor} datasets, 
a great deal of methods have been proposed to tackle the task, which can be broadly divided into depth-only \cite{Uhrig2017THREEDV,chodosh2018deep,ma2018self,eldesokey2019confidence} and multi-sensor fusion based \cite{2018Learning,Cheng2020CSPN,schuster2021ssgp,yan2021dan,gu2021denselidar,yan2021rignet,lin2022dynamic} categories. For example, the literatures \cite{2018Sparse,vangansbeke2019} take sparse depths as the only input to recover dense ones without using color images. Further, Lu \emph{et al.} \cite{2020From} use color images as auxiliary supervision during training and is discarded when testing. Recently, as technology quickly develops, multi-modal information can be captured by sensors, which is beneficial for depth completion. For example, S2D \cite{ma2018self} directly concatenate RGB-D pairs and feed them into networks, contributing to promising improvement. Li \emph{et al.} \cite{li2020multi} propose multi-scale guided cascade hourglass network to handle diverse patterns. PENet \cite{hu2020PENet} proposes to refine depth recovery at three stages. FCFRNet \cite{liu2021fcfr} designs channel-shuffle technology to enhance RGB-D feature fusion. GuideNet \cite{tang2020learning} proposes dynamic convolution to adaptively generate convolution kernels according to color image contents. ACMNet \cite{zhao2021adaptive} conducts graph propagation to extract multi-modal representations. Furthermore, DeepLiDAR \cite{Qiu_2019_CVPR} and PwP \cite{xu2019depth} jointly utilize color images, surface normals, and sparse depths to recover dense depth. FusionNet \cite{vangansbeke2019} and Zhu \emph{et al.} \cite{zhu2021robust} present to estimate uncertainty for robust recovery. NLSPN \cite{park2020nonlocal} and DSPN \cite{xu2020deformable} introduce recurrent non-local and dynamic spatial propagation networks, which significantly improve depth accuracy nearby object boundaries. 

In addition, several unsupervised depth completion works \cite{wong2020unsupervised,krauss2021deterministic,wong2021adaptive,wong2021learning,teutscher2021pdc} also contribute to the development of this domain. For example, KBNet \cite{wong2021unsupervised} proposes the fantastic calibrated backprojection network which achieves very superior performances. However, as mentioned above these methods are designed for dense depth recovery from FoV-limited sparse depth, whilst we aim to learn 360$^{\circ}$ depth completion and 3D reconstruction from panoramic RGB-D input.

\subsection{Monocular Depth Estimation with Full FoV} 
Given panoramic color images, current monocular panoramic depth estimation works mainly devote into predicting 360$^{\circ}$ depths and 3D reconstructions. This topic springs up as soon as the large indoor panoramic datasets Matterport3D \cite{chang2017matterport3d} and Stanford2D3D \cite{armeni2017joint} are constructed in 2017. For this domain in the last five years, supervised methods play a primary role while unsupervised approaches develop slowly. Next we will introduce each of them.

Supervised category: In 2018, OmniDepth \cite{zioulis2018omnidepth} synthesizes 360$^{\circ}$ data with high-quality ground-truth depth annotations by rendering existing datasets. DistConv \cite{tateno2018distortion} proposes distortion-aware convolutional filters to address the inherent distortion of equirectangular projection (EPR) panoramic data. In 2019, Eder \emph{et al.} \cite{eder2019pano} utilize surface normal and plane boundaries to train a plane-aware network to benefit depth estimation. SpherePHD \cite{lee2019spherephd,lee2020spherephd} explores a new data representation via spherical polyhedron, which resolves the shape distortion of spherical panoramas. In 2020, Jin \emph{et al.} \cite{jin2020geometric} and Feng \emph{et al.} \cite{feng2020deep} use geometric priors to help with depth estimation. Wang \emph{et al.} \cite{wang2020bifuse} adopt a two-branch network leveraging EPR and cubemap projections, which are the two most common data forms. In 2021, PanoDepth \cite{li2021panodepth} develops a two-stage framework containing view synthesis and stereo matching. UniFuse \cite{jiang2021unifuse} further improves \cite{wang2020bifuse} with better accuracy and fewer parameters. SliceNet \cite{pintore2021slicenet} transforms the EPR data into slice-based representation, which can tackle the inherent distortion. Sun \emph{et al.} \cite{sun2021hohonet,sun2021indoor} focus on horizontal and vertical contents of a scene for 3D reconstruction. 360MonoDepth \cite{rey2021360monodepth} projects the high-resolution spherical image into tangent image for efficient training. In 2022, SegFuse \cite{feng2022360} utilizes geometric and temporal consistency to constraint depth recovery. GLPanoDepth \cite{bai2022glpanodepth} employs vision transformer and CNNs to encode cubemap and spherical images respectively, obtaining global-to-local representation of panoramas. ACDNet \cite{zhuang2021acdnet} designs adaptively combined dilated convolution to extend receptive field in the EPR and achieves state-of-the-art performances.

Unsupervised category: In 2019, Nikolaos \emph{et al.} \cite{zioulis2019spherical} explore spherical view synthesis for monocular 360$^{\circ}$ depth estimation in a self-supervised manner. In 2021, OlaNet \cite{lai2021olanet} adopts the distortion-aware view synthesis, atrous spatial pyramid pooling, and L1-norm regularized smooth term to effectively and robustly deal with self-supervised panoramic depth estimation. Zhou \emph{et al.} \cite{zhou2021panoramic} combine supervised and unsupervised learning methods to facilitate network training. In 2022, Yun \emph{et al.} \cite{yun2021improving} propose a self-supervised method based on gravity-aligned videos. Similarly, they also utilize the complementarity of supervised and self-supervised learning to improve their models' robustness.

Different from them that only utilize 360$^{\circ}$ color image, our goal is to recover dense depth and 3D reconstruction from the aligned 360$^{\circ}$ color image and sparse depth, which could help improve the accuracy with large margins.

\subsection{Masked Image Encoding for Vision Tasks}
Recently, several Transformer \cite{vaswani2017attention} based approaches \cite{chen2020generative,dosovitskiy2020image,bao2021beit,xie2021simmim,he2021masked} have proved it effective to learn representations from masked images. Specifically, iGPT \cite{chen2020generative} trains a sequence Transformer to auto-regressively predict unknown pixels. ViT \cite{dosovitskiy2020image} conducts masked patch prediction to learn mean color. BEiT \cite{bao2021beit} presents to predict tokenization. SimMIM \cite{xie2021simmim} and MAE \cite{he2021masked} propose to recover raw pixels of randomly masked patches by a lightweight one-layer head and an asymmetric decoder, respectively. In contrast to them, our M$^{3}$PT is technically designed for multi-modal vision tasks instead of the single-modal image-based recognition.

\section{Method}
In this section, we first introduce how to synthesize sparse depth data and then elaborate on the multi-modal masked pre-training strategy.

\subsection{Data Synthesis}

{All existing panoramic datasets do not provide sparse depth for 360$^\circ$ depth completion task. However, the sparse depth data can be possibly captured by some actual products such as Matterport Pro2 and FARO Focus 3D cameras. Limited by the lack of these hardware devices, in this paper, we imitate the principle of laser scanning to produce 360$^\circ$ sparse depth sampled from the dense ground-truth depth annotation, aiming at synthesizing the sparse depth data that matches the actual products as much as possible}. The sampling principle is similar to that of KITTI benchmark \cite{Uhrig2017THREEDV} which provides depth with about 7\% density captured by 64-line LiDAR scanning.


 \begin{figure}[t]
  \centering
  \includegraphics[width=0.98\columnwidth]{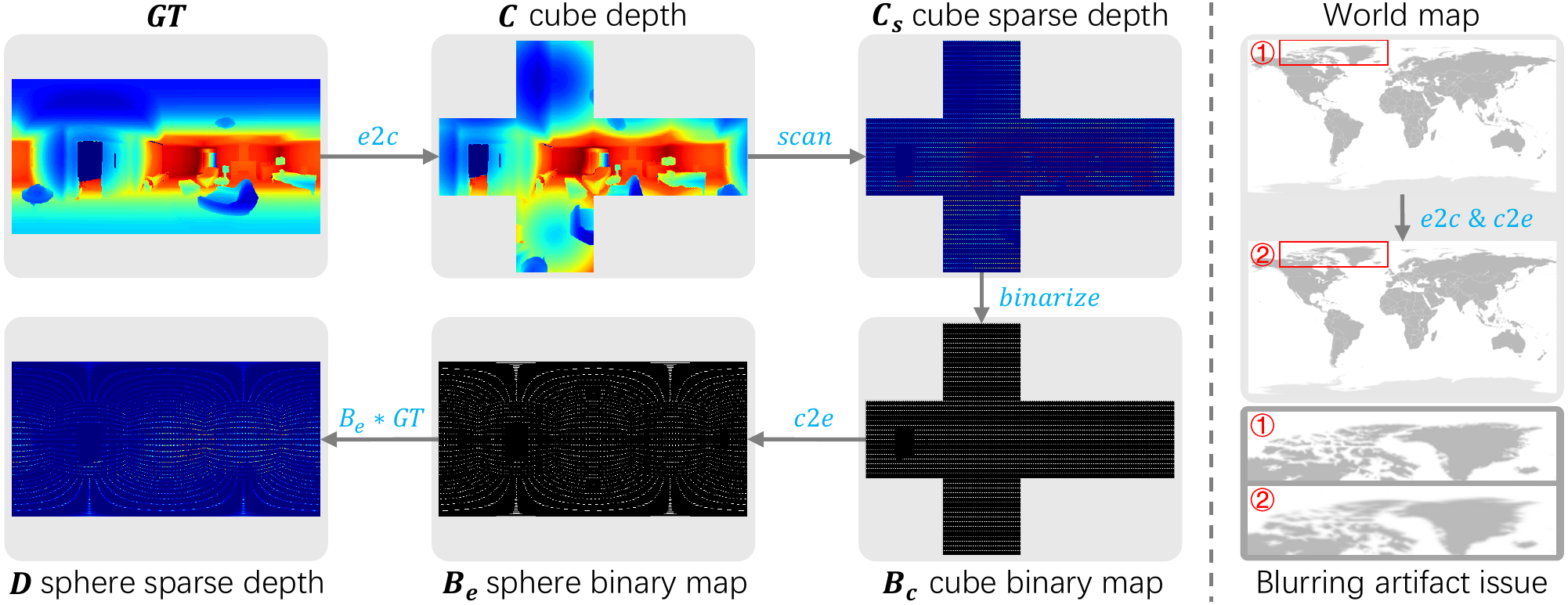}\\
  \caption{A flowchart of data synthesis (left) and an example of the blurring artifact issue (right) that has negative effects on data processing. Best color of view.}\label{data_syn}
\end{figure}

As illustrated in the left of Figure \ref{data_syn}, the ground-truth depth $\boldsymbol{GT}$ is stored in spherical view by equirectangular projection, which brings inherent distortion. Hence, it's inaccurate 
to produce sparse depth directly based on $\boldsymbol{GT}$ in scanning mode. As an alternative, we first project the equirectangular $\boldsymbol{GT}$ into cubical map $\boldsymbol{C}$ by \emph{e2c} function, ignoring the inherent distortion. Next, we generate cube sparse depth $\boldsymbol{C_s}$ via imitating the laser scanning, \emph{e.g.}, taking one pixel for every eight pixels horizontally and one pixel for every two pixels vertically. Then $\boldsymbol{C_s}$ is binarized to obtain cube binary map $\boldsymbol{B_c}$. $\boldsymbol{B_c}$ is thus converted into $\boldsymbol{B_e}$ by \emph{c2e} function. Finally, we acquire the desired sparse depth $\boldsymbol{D}$ multiply $\boldsymbol{B_e}$ by $\boldsymbol{GT}$. The process can be simply defined as: 
\begin{equation}\label{eq_data_syn}
\begin{split}
\boldsymbol{D}=\boldsymbol{B_e}*\boldsymbol{GT},
\end{split}
\end{equation}
where $\boldsymbol{B_e}=f\left( \boldsymbol{B_c|\boldsymbol{C_s}, \boldsymbol{C}, \boldsymbol{GT}} \right)$, $f\left( \cdot \right)$ refers to the combination of \emph{e2c}, \emph{scan}, \emph{binarize}, and \emph{c2e}. {The details of \emph{e2c} and \emph{c2e} functions refer to this project\footnote{https://github.com/sunset1995/py360convert}.}

Note that, it is theoretically possible to use \emph{c2e} to directly project the cubical $\boldsymbol{C_s}$ into the equirectangular $\boldsymbol{D}$. However, this would lead to blurring artifacts in the polar region, as evidenced in the right part of Figure \ref{data_syn}. Instead, we choose to project a binary map and then use it to accurately sample valid points from $\boldsymbol{GT}$. In this way, our method can reduce error pixels as much as possible.

\subsection{Multi-Modal Masked Pre-Training}
\begin{figure}[t]
  \centering
  \includegraphics[width=1\columnwidth]{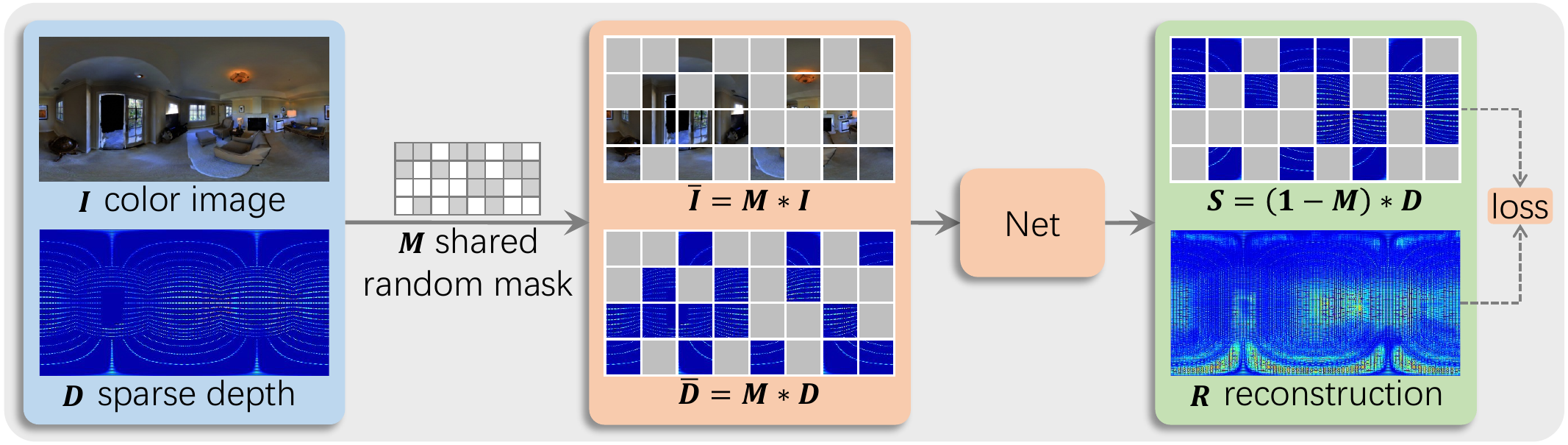}\\
  \caption{\textbf{Our M$^{3}$PT pipeline}. During pre-training, the input color image $\boldsymbol{I}$ and sparse depth $\boldsymbol{D}$ are masked out by the shared random mask $\boldsymbol{M}$, obtaining $\boldsymbol{\bar{I}}$ and $\boldsymbol{\bar{D}}$ respectively. Then $\boldsymbol{\bar{I}}$ and $\boldsymbol{\bar{D}}$ are fed into a network to predict the depth reconstruction $\boldsymbol{R}$, supervised by the signal $\boldsymbol{S}$ which is the complementary set of $\boldsymbol{\bar{D}}$ in $\boldsymbol{D}$. After pre-training, with $\boldsymbol{I}$ and $\boldsymbol{D}$ as input, the network with learned initial weights is applied to recover target depths supervised by dense ground-truth depth annotations.}\label{pipeline}
\end{figure}

As shown in Figure \ref{pipeline}, our multi-modal masked pre-training (M$^{3}$PT) for 360$^{\circ}$ depth completion is a simple strategy that reconstructs the sparse depth signal given partial observations of the RGB-D pair under shared random mask. Here we introduce the key components of M$^{3}$PT and explicitly analyze the differences between recent visual masked pre-training approaches (\eg, MAE \cite{he2021masked}, SimMIM~\cite{xie2021simmim}) and M$^{3}$PT.

\textbf{Shared random mask.} 
Different from MAE where masking is performed on single RGB data, 
we propose to mask both RGB image and sparse depth with the \emph{shared random mask} to produce incomplete RGB-D pair as input for pre-training. In fact, there are other options of masking strategies for PDC task, including (i) only mask the RGB image, (ii) only mask the sparse depth, and (iii) mask both RGB image and sparse depth but with different random masks. Unfortunately, all of these strategies have a risk of leaking information from another modality, preventing the pre-training task from learning robust semantics based on the multi-modal context. We will show the comparisons between different masking strategies later in the experimental part.

\textbf{Backbone.} Our method is flexible and can be applied to any existing approach which receives the RGB-D pair as input. In this paper, we mainly choose GuideNet \cite{tang2020learning} as backbone for the majority of our experiments. In addition, we also test the effectiveness of M$^3$PT using UniFuse \cite{jiang2021unifuse} and HoHoNet \cite{sun2021hohonet}. Note that there is no need to design extra modules (e.g., a decoder) for the architectures of these existing approaches, even when they have an additional pre-training stage in M$^3$PT. It is because that the regression targets are physically similar between pre-training and fine-tuning stages. See more details in the `Reconstruction target' part as follows.

\textbf{Reconstruction target.} The reconstruction target of M$^{3}$PT is the sparse depth on the masked regions. It is quite different from the popular masked pre-training methods~\cite{he2021masked,xie2021simmim} in vision where the missing image pixels are predicted. Compared to the vision pre-training counterparts~\cite{he2021masked,xie2021simmim}, this design has two obvious advantages: (i) it closes the gap between pre-training and fine-tuning tasks, as they differ only in the prediction density; (ii) it leads to \emph{no architectural modification} between pre-training and fine-tuning stages, which can potentially make the transfer learning more smooth and effective.

\section{Experiments}
Here, we first report datasets and metrics. Next, extensive ablation studies are conducted to verify the effectiveness of the proposed M$^3$PT. Then, we compare against other state-of-the-art (SoTA) works on three datasets. At last, we validate the generalization capability of M$^3$PT on KITTI benchmark \cite{Uhrig2017THREEDV}.  

\subsection{Datasets}



We conduct our experiments on three commonly used benchmark datasets of real world, \ie, Matterport3D\footnote{https://vcl3d.github.io/Pano3D/download/} \cite{albanis2021pano3d}, Stanford2D3D\footnote{http://buildingparser.stanford.edu/dataset.html} \cite{armeni2017joint}, and 3D60\footnote{https://vcl3d.github.io/3D60/} \cite{zioulis2019spherical}. Matterport3D is a scanned dataset collected by Matterport's Pro 3D camera. The latest Matterport3D ($512\times 256$) consists of 7,907 panoramic RGB-D pairs, of which 5636 for training, 744 for validating, and 1527 for testing. Stanford2D3D is composed of 1,413 panoramic color images and corresponding depth maps, whose training and testing splits contain 1,040 and 373 RGB-D pairs, respectively. We resize them to $512\times 256$. 3D60 is initially made up of Matterport3D, Stanford2D3D, and SunCG \cite{song2017semantic}. But now it skips the entire SunCG dataset considering legal matters. As a result, the latest 3D60 ($512\times 256$) consists of 6,669 RGB-D pairs for training, 906 for validating, and 1831 for testing, 9,406 in total.

\subsection{Metrics}
Following previous works \cite{sun2021hohonet,pintore2021slicenet,yun2021improving,zhuang2021acdnet}, we use five common and standard metrics to evaluate our methods, including MRE, MAE, RMSE, RMSElog, and ${{\delta }_{i}}$ ($i=1.25, 1.25^2, 1.25^3$). Please refer to our appendix for more details.

\begin{table}[t]
\centering
\renewcommand\arraystretch{1.2}
\resizebox{0.7\textwidth}{!}{
\begin{tabular}{c|c|c|cccc}
\Xhline{1.2pt}
\multirow{2}{*}{Masked Data} & \multirow{2}{*}{Shared Mask} & \multirow{2}{*}{Mask Ratio}  & \multicolumn{4}{c}{Mask Size} \\ \cline{4-7} 
                             &                              &             & 4     & 8     & 16     & 32     \\ \hline
RGB                          & -                            & \multirow{4}{*}{0.75}            & 195.7  & 198.1 & 194.0     & 196.2    \\
D                            & -                            &             & 190.7 & 177.6 & 186.1  & 203.3  \\
RGB-D                        & No                           &             & 183.4 &  178.5 & 182.2 &    193.0    \\
RGB-D                        & Yes                          &             & \textbf{166.8} & \textbf{168.9} & \textbf{167.6}  & \textbf{169.9}  \\ 
\Xhline{1.2pt}
\end{tabular}
}
\caption{Ablation on different masked input data on Stanford2D3D dataset, where the metric is RMSE (mm). The error of baseline without pre-training is \textbf{196.7}.}
\label{ab_masked_data}
\end{table}

\begin{table}[t]
\centering
\large
\renewcommand\arraystretch{1.2}
\resizebox{1\textwidth}{!}{
\begin{tabular}{l|ccc|ccc|cccccc|ccc}
\Xhline{1.2pt}
Mask Size  & \multicolumn{3}{c|}{4} & \multicolumn{3}{c|}{8} & \multicolumn{6}{c|}{16}              & \multicolumn{3}{c}{32} \\ \hline
Mask Ratio & 0.45   & 0.6   & 0.75  & 0.45   & 0.6   & 0.75  & 0.15  & 0.3   & 0.45  & 0.6   & 0.75 & 0.9 & 0.45   & 0.6   & 0.75 \\
RMSE       & 172.4  & 170.3 & 168.8 & 171.2  & 169.3 & 168.9 & 173.5 & 169.4 & 168.9 & 169.7 & \textbf{167.6} & 169.3 & 173.7  & 172.4  & 169.9 \\ 
\Xhline{1.2pt}
\end{tabular}
}
\caption{Ablation on different mask sizes and mask ratios on Stanford2D3D dataset.}
\label{ab_mask_size_ratio}
\end{table}

\subsection{Ablation Studies}
\textbf{Settings}: 
{We employ GuideNet \cite{tang2020learning} as the default backbone.} The model is pre-trained for 300 epochs and fine-tuned for 100 epochs on every dataset. The mask is randomly generated following \cite{xie2021simmim,he2021masked} with different sizes and ratios.

\noindent\textbf{(1) Masking strategy.} 

\textbf{(i)} We explore how to corrupt RGB-D data during pre-training in Table \ref{ab_masked_data}. We can find that shielding only RGB, only Depth, or RGB-D without shared mask, all of which lead to worse performances because these operations destroy the model's learning of unknown areas. In contrast, the model achieves the best results when employing the proposed shared random mask, indicating that corrupting the same areas can contribute to improvement for multi-modal vision tasks. The following experiments are based on the random shared mask.

\textbf{(ii)}  We study the effect of different mask sizes and ratios on the model's representation learning in Table \ref{ab_mask_size_ratio}. First, the model performs better when the mask size is changed from 4 to 16. We hold the opinion that the larger mask urges the model to learn long-range dependency between invisible and visible pixels. However, when setting the mask size to 32, the model has a degraded performance as it is too large to establish remote dependency. Second, when the mask size is set to 16, the model tends to perform better from 15\% to 90\% ratios, which could enforce the model to predict more unseen areas and acquires representation that is closer to the real domain.

\noindent\textbf{(2) Number of pre-training epochs and data amounts.} 

\begin{figure}[t]
\centering
\includegraphics[width=0.99\columnwidth]{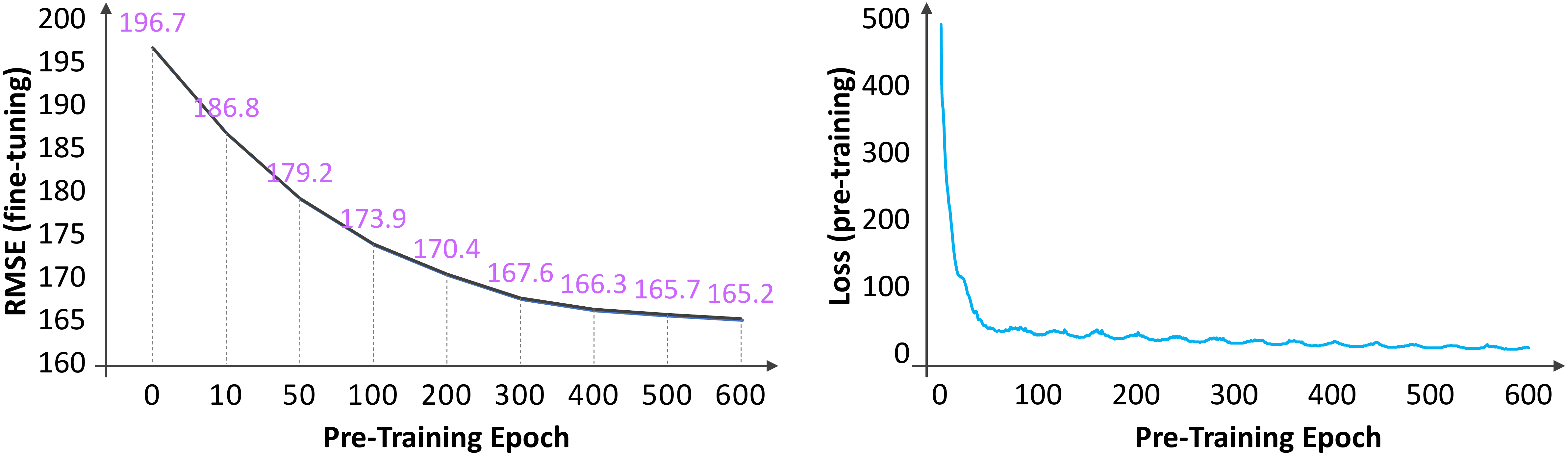}\\
\caption{Ablation on different pre-training epochs with mask size 16 and mask ratio 0.75 on Stanford2D3D dataset. The loss value is magnified by 10$^{4}$ for clear visualization.}\label{ablation_epoch}
\end{figure}

\textbf{(i)} The left of Figure \ref{ablation_epoch} demonstrates the influence of different pre-training epochs on fine-tuning, and the right shows the loss of pre-training. We can find that the model's error gradually decreases with the increase of pre-training epochs. This is because the model can learn better representation with more epochs, which is also reflected in the lower loss in the right of Figure \ref{ablation_epoch}. For the trade-off between speed and accuracy, unless otherwise stated, we pre-train the model for 300 epochs by default. 

\textbf{(ii)} Table \ref{ab_number_of_dataset_for_pre-training} explores the  effect of different data amounts for pre-training on fine-tuning. 
For a fair comparison, we report the results of the 400th epoch without pre-training, whose cost roughly aligns with the setting of pre-training for 300 epochs and fine-tuning for 100 epochs. It can be found that without pre-training, the performance of 400 epochs has no improvements over 100 epochs. However, when conducting M$^3$PT just on single dataset, it leads to 12.9\% improvements averagely on three datasets, demonstrating the significant effectiveness of M$^3$PT. Further, when pre-training on all the data of these three datasets, the performances are always superior to that only using a single dataset. Therefore, it is concluded that more data involved in M$^{3}$PT can consistently prevents the overfitting risks during fine-tuning.

\begin{table}[t]
\centering
\renewcommand\arraystretch{1.2}
\resizebox{0.98\textwidth}{!}{
\begin{tabular}{l|c|c|c|c|c}
\Xhline{1.2pt}
\multirow{2}{*}{Dataset} & Data  & w/o Pret. & w/o Pret. & Pret. on Self Data & Pret. on All Data\\
               & amount& (100 epochs) & (400 epochs) & (300+100 epochs) & (300+100 epochs) \\ \hline
Matterport3D   & 5.6k  & 168.1  & 168.5   & 146.2            & \textbf{138.9} \\
Stanford2D3D   & 1k    & 196.7  & 196.3   & 167.6            & \textbf{149.0} \\
3D60           & 6.7k  & 159.9  & 160.2   & 142.3            & \textbf{127.2} \\ \Xhline{1.2pt}
\end{tabular}
}
\caption{Ablation on different data amounts used during pre-training (Pret.). The mask size and ratio are 16 and 0.75, respectively. ``300+100 epochs'' denotes 300 pre-training epochs and 100 fine-tuning epochs.}
\label{ab_number_of_dataset_for_pre-training}
\end{table}

\begin{table}[t]
\centering
\renewcommand\arraystretch{1.2}
\resizebox{0.96\textwidth}{!}{
\begin{tabular}{c|c|p{1.1cm}<{\centering}p{1.1cm}<{\centering}p{1.1cm}<{\centering}c|p{1.1cm}<{\centering}p{1.1cm}<{\centering}p{1.1cm}<{\centering}}
\Xhline{1.2pt}
\multirow{2}{*}{Dataset}  & \multirow{2}{*}{Method}  & \multicolumn{4}{c|}{Error Metric $\downarrow$}  & \multicolumn{3}{c}{Accuracy Metric $\uparrow$} \\ \cline{3-9}

&  & MRE & MAE & RMSE & RMSElog  & $\boldsymbol{{\delta }_{1.25}}$  & $\boldsymbol{{\delta }_{{1.25}^{2}}}$ & $\boldsymbol{{\delta }_{{1.25}^{3}}}$ \\ \hline
\multirow{6}{*}{\begin{sideways}{Matterport3D}\end{sideways}} 
& UniFuse \cite{jiang2021unifuse}  & 0.0475 & 95.2  & 229.1 & 0.0381 & 0.9710 & 0.9924 & 0.9970 \\
& HoHo-R \cite{sun2021hohonet}     & \underline{0.0355} & \underline{75.0}  & 199.2 & \underline{0.0311} & \underline{0.9806} & 0.9945 & 0.9977 \\
& HoHo-H \cite{sun2021hohonet}     & 0.0406 & 85.7  & 215.5 & 0.0337 & 0.9772 & 0.9938 & 0.9975 \\
& PENet \cite{hu2020PENet}         & 0.0493 & 91.5  & 248.0 & 0.0350 & 0.9728 & 0.9935 & 0.9970 \\
& GuideNet \cite{tang2020learning} & 0.0438 & 87.2  & \underline{192.9} & 0.0327 & \underline{0.9806} & \underline{0.9948} & \underline{0.9981} \\
& \textbf{M$^3$PT}             & \textbf{0.0164} & \textbf{36.2}  & \textbf{138.9} & \textbf{0.0193} & \textbf{0.9927} & \textbf{0.9976} & \textbf{0.9990} \\ \hline
\multirow{6}{*}{\begin{sideways}{Stanford2D3D}\end{sideways}}
& UniFuse \cite{jiang2021unifuse}  & \underline{0.0489} & 93.4  & 216.2 & 0.0392 & 0.9661 & 0.9919 & 0.9973 \\
& HoHo-R \cite{sun2021hohonet}     & 0.0677 & 123.9 & 242.5 & 0.0478 & 0.9463 & 0.9862 & 0.9959 \\
& HoHo-H \cite{sun2021hohonet}     & 0.0695 & 127.9 & 254.8 & 0.0497 & 0.9434 & 0.9852 & 0.9957 \\
& PENet \cite{hu2020PENet}         & 0.0530 & 95.9  & 200.6 & 0.0404 & \underline{0.9694} & \underline{0.9934} & \underline{0.9981} \\
& GuideNet \cite{hu2020PENet}      & 0.0506 & \underline{92.1}  & \underline{196.7} & \underline{0.0380} & 0.9689 & 0.9926 & 0.9978 \\
&\textbf{M$^3$PT}              & \textbf{0.0274} & \textbf{52.9}  & \textbf{149.0} & \textbf{0.0263} & \textbf{0.9859} & \textbf{0.9963} & \textbf{0.9988} \\ \hline
\multirow{6}{*}{\begin{sideways}{3D60}\end{sideways}}        
& UniFuse \cite{jiang2021unifuse}  & 0.0446 & 94.1  & 215.6 & 0.0342 & 0.9749 & 0.9947 & 0.9984 \\
& HoHo-R \cite{sun2021hohonet}     & \underline{0.0338} & \underline{75.6}  & \underline{196.9} & \underline{0.0294} & \underline{0.9818} & \underline{0.9954} & 0.9983 \\
& HoHo-H \cite{sun2021hohonet}     & 0.0376 & 81.9  & 205.8 & 0.0317 & 0.9788 & 0.9947 & 0.9981 \\
& PENet \cite{hu2020PENet}         & 0.0680 & 120.3 & 233.9 & 0.0321 & 0.9743 & 0.9926 & 0.9980 \\
& GuideNet \cite{hu2020PENet}      & 0.0689 & 144.2 & 239.3 & 0.0418 & 0.9711 & \underline{0.9954} & \underline{0.9987} \\
& \textbf{M$^3$PT}             & \textbf{0.0144} & \textbf{34.1} & \textbf{127.2} & \textbf{0.0165} & \textbf{0.9944} & \textbf{0.9985} & \textbf{0.9995} \\ \Xhline{1.2pt}
\end{tabular}
}
\caption{Quantitative comparisons of panoramic depth completion on three datasets. The best and the second best results are highlighted in bold and underline, respectively.}
\label{compare_with_sota}
\end{table}

 \begin{figure}[t]
  \centering
  \includegraphics[width=1\columnwidth]{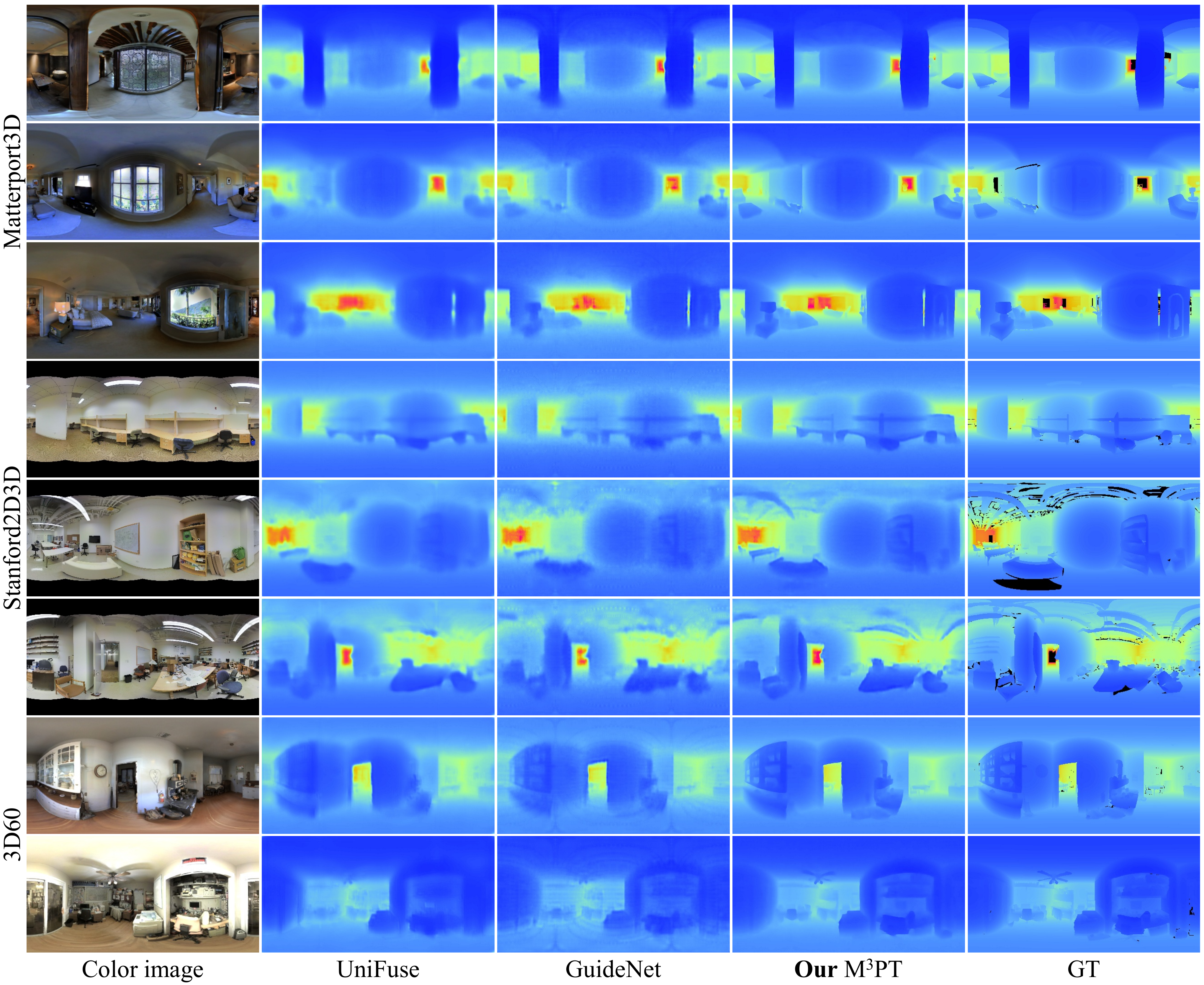}\\
  \caption{Qualitative comparison of different methods, including UniFuse \cite{jiang2021unifuse}, GuideNet \cite{tang2020learning}, and our M$^3$PT. More visualizations can be found in our supplementary material.}\label{vis_compare_with_sota}
\end{figure}

\subsection{Comparisons with SoTA Methods}
\begin{figure}[t]
  \centering
  \includegraphics[width=1\columnwidth]{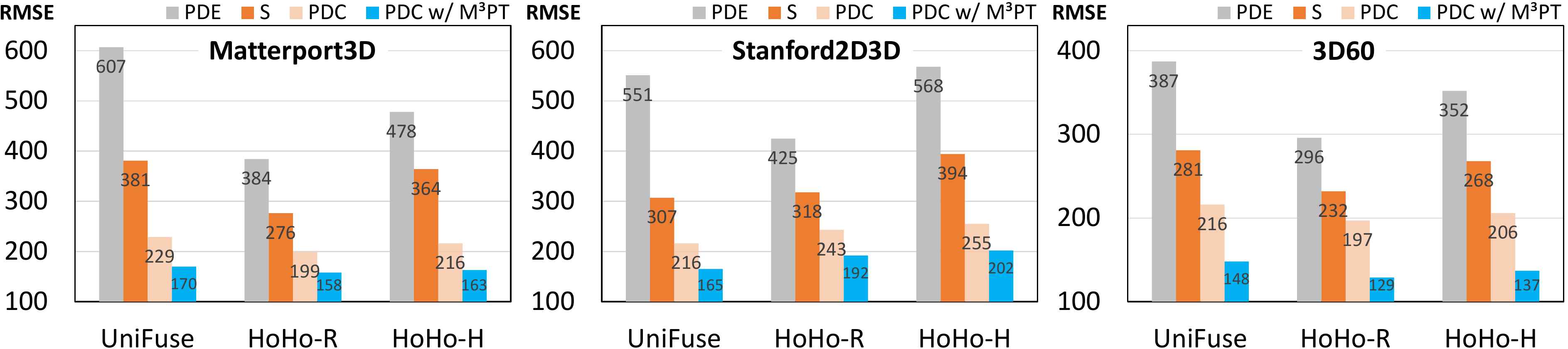}\\
  \caption{Comparisons of different baselines with different-modal input data and M$^3$PT. PDE: panoramic depth estimation only from color images. PDC: panoramic depth completion from not only color images but the corresponding sparse depths.
  }\label{comparison_de_dc_M3PT}
\end{figure}

In this subsection, we compare with recent SoTA works, including UniFuse \cite{jiang2021unifuse}, HoHoNet \cite{sun2021hohonet}, PENet \cite{hu2020PENet}, and GuideNet \cite{tang2020learning}. HoHo-R and HoHo-H severally refer to using ResNet \cite{He2016Deep} and HardNet \cite{chao2019hardnet} as its backbone. Table \ref{compare_with_sota} and Figure \ref{vis_compare_with_sota} demonstrate the quantitative and qualitative results, respectively. Based on different baselines, Figure \ref{comparison_de_dc_M3PT} further shows the influence of additional sparse depth information and proposed M$^{3}$PT on the recovery of panoramic depth.

\noindent\textbf{(1) Quantitative results.} 

Overall, as illustrated in Table \ref{compare_with_sota}, the proposed M$^{3}$PT is consistently superior to other methods in all metrics on three datasets.

\textbf{(i)} On Matterport3D dataset, M$^{3}$PT greatly exceeds the second-best HoHo-R by 53.8\%, 51.7\%, and 37.9\% in MRE, MAE, and RMSElog, severally. Compared with the suboptimal GuideNet in RMSE, the error is reduced from 192.9mm to 138.9mm, improving the performance nearly by 28.0\%. Besides, M$^{3}$PT achieves the highest accuracies in ${{\delta }_{i}}$ with different thresholds, outperforming the second-best method by 1.21, 0.29, and 0.09 percent point in ${{\delta }_{1}}$, ${{\delta }_{2}}$, and ${{\delta }_{3}}$, respectively.

\textbf{(ii)} On Stanford2D3D dataset, M$^{3}$PT is superior to the suboptimal UniFuse with 44.0\% improvement in MRE. Also, the MAE, RMSE, RMSElog is severally reduced by 42.6\%, 24.3\%, and 30.8\% when comparing M$^{3}$PT with the second-best GuideNet. In addition, the accuracy metric verifies the effectiveness of M$^{3}$PT again, which plays a prominent role in all approaches.

\textbf{(iii)} On 3D60 dataset, M$^{3}$PT surpasses the second-best HoHo-R with large margins, improving it by 57.4\% in MRE, 54.9\% in MAE, 35.4\% in RMSE, and 43.9\% in RMSElog, severally. Furthermore, M$^{3}$PT is more accurate than other approaches and prevail over the suboptimal methods with 1.26, 0.31, and 0.08 percent point in ${{\delta }_{1}}$, ${{\delta }_{2}}$, and ${{\delta }_{3}}$, respectively.

\textbf{(iv)} Last but not least, apart from GuideNet that has been reported in Table \ref{compare_with_sota}, we further employ UniFuse, HoHo-R, and HoHo-H as baselines to see the influence of the additional sparse depth data and proposed mask strategy on the recovery of panoramic depth. As shown in Figure \ref{comparison_de_dc_M3PT}, gray bar: only using RGB, orange bar: only using sparse depth, light orange bar: using both RGB and its sparse depth, and blue bar: using RGB and sparse depth with the proposed mask strategy M$^{3}$PT. We can find that the error of only using sparse depth is much lower than that of only using RGB. Also, adding sparse depth data can benefit models with very large margins. Specifically, comparing light orange bar with gray bar, the errors of UniFuse, HoHo-R, and HoHo-H are severally reduced by 55.8\%, 41.5\%, and 50.5\% on average on three datasets. What's more, adopting M$^{3}$PT contributes to their significant improvements of 27.0\%, 25.4\%, and 26.3\% on average compared with the orange bars on three datasets. These facts indicate sparse depth information has great reference value for depth recovery, and also prove that the panoramic depth completion is a potentially valuable task.


 \begin{figure}[t]
  \centering
  \includegraphics[width=1\columnwidth]{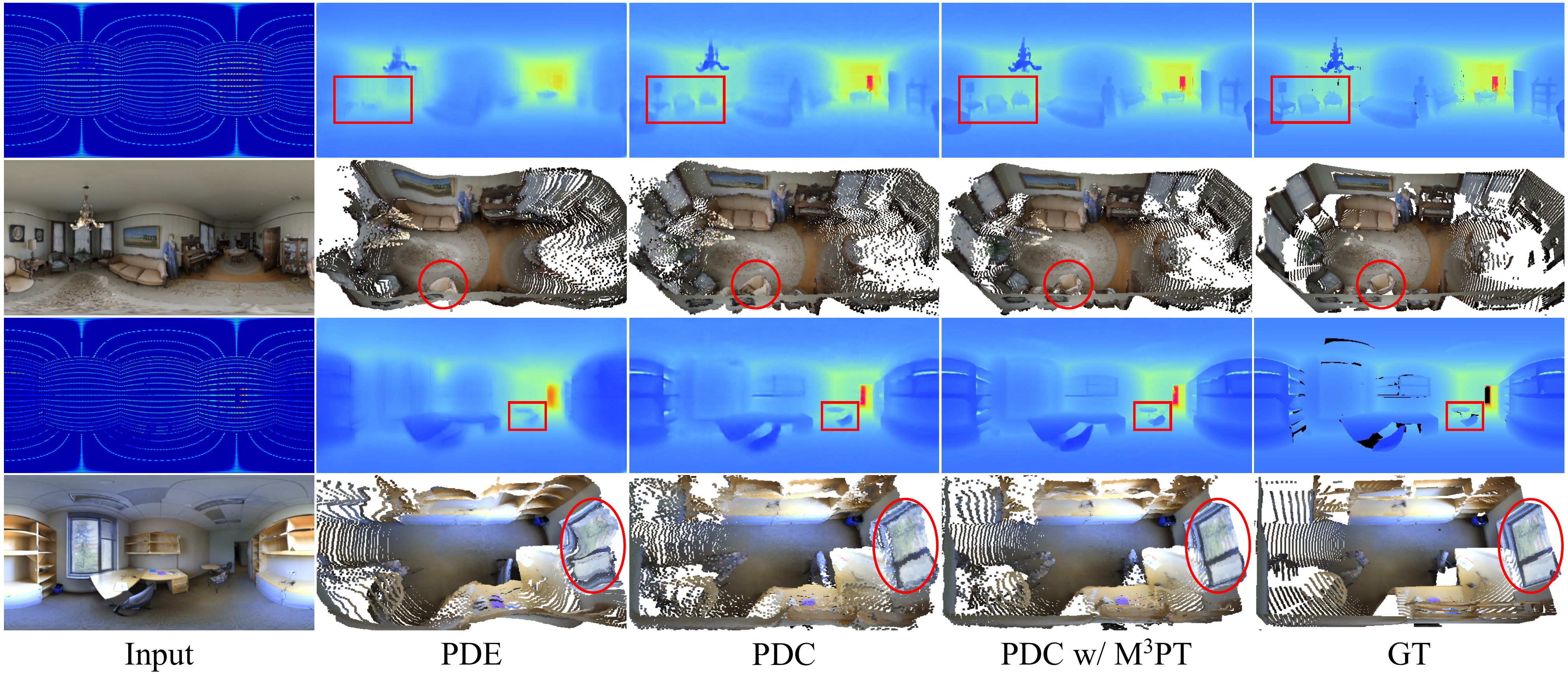}\\
  \caption{Visual results of UniFuse \cite{jiang2021unifuse} with different-modal input data and M$^3$PT.}\label{vis_different_task}
\end{figure}

\noindent\textbf{(2) Qualitative results.} 

\textbf{(i)} As shown in Figure \ref{vis_compare_with_sota}, our M$^{3}$PT can recover more detailed objects and precise depth with reasonable visual effect. For example, for one thing, as illustrated in the first, fourth, and seventh rows, M$^{3}$PT succeeds in predicting clearer edges of doors, tables, chairs, windows, \emph{etc}. For another thing, as shown in the fifth and sixth rows, although the color images of Stanford2D3D do not have pixels at both the top and bottom, M$^{3}$PT can still predict more accurate depth values in the invisible areas. It strongly demonstrates the effectiveness and generalization of the proposed masked pre-training strategy via corrupting RGB-D data. In addition, M$^{3}$PT is also good at distinguishing from background and foreground, \emph{e.g.}, the furniture can be clearly discriminated from the wall.

\textbf{(ii)} As illustrated in Figure \ref{vis_different_task}, based on only color images (PDE), the depth predicted by UniFuse is extremely blurry that the corresponding 3D reconstruction introduces plenty of wrong location information, which causes negative deformation, especially nearby walls. By contrast, adding sparse depth data (PDC) vastly improves the visual effect of both depth recovery and 3D reconstruction. Furthermore, when deploying the proposed M$^{3}$PT with RGB-D data as input, both objects' structures and details tend to be more clear and abundant.


\subsection{Generalization Capability}
In this subsection, we further verify the generalization capability of M$^3$PT on \href{http://www.cvlibs.net/datasets/kitti/eval_depth.php?benchmark=depth_completion}{KITTI depth completion benchmark}, whose sparse depth data is obtained by a 64-line LiDAR, and the RGB-D pairs have limited field of vision. As reported in Table \ref{gc_kitti}, M$^3$PT consistently improves the performances of S2D, GuideNet, and ACMNet. For example, M$^3$PT reduces RMSE/MAE by 15.05mm/18.36mm averagely, indicating that our M$^3$PT possesses robust generalization capability.

\begin{table}[t]
\centering

\renewcommand\arraystretch{1.2}
\resizebox{0.55\textwidth}{!}{
\begin{tabular}{l|cccc}
\Xhline{1.2pt}
Method    & RMSE  & MAE  & iRMSE  & iMAE \\ \hline
S2D \cite{ma2018self}  
& 858.02    & 311.47   & 3.07    & 1.67 \\
+M$^3$PT  
& \textbf{844.16}  & \textbf{267.64}  & \textbf{3.01} &  \textbf{1.51} \\ \hline
GuideNet \cite{tang2020learning}  
& 777.78    & 221.59   & 2.39    & \textbf{1.00}  \\
+M$^3$PT  
& \textbf{761.57}  & \textbf{217.68}  & \textbf{2.26}  & \textbf{1.00}  \\ \hline
ACMNet \cite{zhao2021adaptive}   
& 789.72  & 216.65  & 2.32  & 0.96  \\
+M$^3$PT  
& \textbf{774.63}  & \textbf{209.31}  & \textbf{2.25}  & \textbf{0.93}  \\ \Xhline{1.2pt}
\end{tabular}
}
\caption{Performances of M$^3$PT with different baselines on KITTI validation split.}
\label{gc_kitti}
\end{table}

\section{Conclusion}
In this paper, we introduced a potentially valuable task, \emph{i.e.}, panoramic depth completion, to help with dense panoramic depth recovery and 3D reconstruction from monocular 360$^\circ$ RGB-D data. Furthermore, we proposed the multi-modal masked pre-training (M$^3$PT) framework to handle this task. It was the first time we showed that the masked pre-training could be very effective in modeling multi-modal tasks for vision, instead of the single-modal image recognition which was popularized by the masked autoencoders (MAE). As a result, comprehensive evaluations demonstrated the superiority of M$^3$PT on three benchmark datasets. At last, we hope our exploration in this paper can facilitate future studies concerned with multi-modal vision tasks. In the future, we are going to extend M$^3$PT to related topics such as depth denoising and super-resolution.

\section{Acknowledgement}
The authors would like to thank reviewers for their detailed comments and instructive suggestions. This work was supported by the National Science Fund of China under Grant Nos. U1713208, 62072242 and Postdoctoral Innovative Talent Support Program of China under Grant BX20200168, 2020M681608. Note that the PCA Lab is associated with, Key Lab of Intelligent Perception and Systems for High-Dimensional Information of Ministry of Education, and Jiangsu Key Lab of Image and Video Understanding for Social Security, Nanjing University of Science and Technology.

\bibliographystyle{splncs04}
\bibliography{egbib}

\clearpage

\end{document}